\documentclass[10pt, conference]{IEEEtran}
\usepackage{cite}
\usepackage{amsmath,amssymb,amsfonts}
\usepackage{graphicx}
\usepackage{textcomp}
\usepackage{xcolor}
\usepackage{bm}
\usepackage{cases}
\usepackage[skip=1pt, belowskip=0pt]{caption}
\usepackage{subfig}
\usepackage[ruled,linesnumbered,noend]{algorithm2e} 
\usepackage{setspace} % used to stretch out linespacing in algorithm2e
\usepackage[makeroom]{cancel} % for crossing out math, in particular the $\in$ command
\usepackage{amsthm}
\usepackage{enumitem}  

\usepackage[normalem]{ulem} % for strikeout

    \usepackage{physics}
    \usepackage{amsmath}
    \usepackage{tikz}
    \usepackage{mathdots}
    \usepackage{yhmath}
    \usepackage{cancel}
    \usepackage{color}
    \usepackage{siunitx}
    \usepackage{array}
    \usepackage{multirow}
    \usepackage{amssymb}
    \usepackage{gensymb}
    \usepackage{tabularx}
    \usepackage{booktabs}
    \usetikzlibrary{patterns}
    \usetikzlibrary{shapes}
    
\makeatletter
\def\BibTeX{{\rm B\kern-.05em{\sc i\kern-.025em b}\kern-.08em
    T\kern-.1667em\lower.7ex\hbox{E}\kern-.125emX}}
    
\allowdisplaybreaks
    
\newcommand{\Rbb}{\mathbb{R}}
\newcommand{\inrbb}[1]{\in\Rbb^{#1}}

\newcommand{\Nbb}{\mathbb{N}}
 
\mathchardef\myhyphen="2D

\newcommand{\Amat}{\matr{A}}

\newcommand{\task}[1]{(\Amat_{#1},\,\yvec_{#1})}
\newcommand{\wstar}{\wvec_\star}
\newcommand{\Atau}[1]{\Amat_{\tau(#1)}}
\newcommand{\Atauk}[2]{\Amat_{\tau(#1),[#2]}}
\newcommand{\ytau}[1]{\yvec_{\tau(#1)}}
\newcommand{\tasktau}[1]{(\Atau{#1},\,\ytau{#1})}
\newcommand{\ntau}[1]{n_{\tau({#1})}}
\newcommand{\Abar}{\bar{\Amat}}
\newcommand{\Abartau}[1]{\Abar_{\tau(#1)}}
\newcommand{\Ptau}[1]{\Pmat_{\tau(#1)}}
\newcommand{\Abb}{\Amat_{\Sc}}
\newcommand{\Yvec}{\yvec_{\Sc}}
\newcommand{\weven}{\wvec_{\text even}}
\newcommand{\wodd}{\wvec_{\text odd}}

\newcommand{\cocoa}{\mbox{\textsc{CoCoA}}}

\newcommand{\dxk}{\Delta \xvec_{[k]}}
\newcommand{\dxki}{\dxk^i}

\newcommand{\subproblemp}{\sigma'}
\newcommand{\aggregationp}{\gamma}

\newcommand{\matr}[1]{\bm{#1}}

\newcommand{\Pmat}{\matr{P}}

\newcommand{\vvec}{\matr{v}}
\newcommand{\vbar}{\bar{\matr{v}}}
\newcommand{\wvec}{\matr{w}}
\newcommand{\xvec}{\matr{x}}

\newcommand{\yvec}{\matr{y}}

\newcommand{\deltavec}{\matr{\delta}}

\newcommand{\Imat}{\matr{I}}

\newcommand{\eye}[1]{\Imat_{#1}}

\DeclareMathAlphabet{\mymathbb}{U}{BOONDOX-ds}{m}{n}

\newcommand{\T}{^\intercal} % Transpose
\newcommand{\p}{^+}

\let\oldin\in
\renewcommand{\in}{{\,\oldin\,}}

\let\oldnotin\notin
\renewcommand{\notin}{{\,\oldnotin\,}}

\newcommand{\krange}{{k=1,\,\dots,\,K}}

\long\def\/*#1*/{}

\newcommand{\Fc }{\mathcal{F}}

\newcommand{\Pc}{\mathcal{P}}

\newcommand{\Sc }{\mathcal{S}}

\definecolor{myblue}{rgb}{0.3328, 0.3539, 0.7758}
\definecolor{myblue2}{rgb}{0.0328, 0.0539, 0.4758}
\definecolor{mygreen2}{rgb}{ 0.0328 0.4758 0.0539} 
\definecolor{mygreen3}{rgb}{ 0.0328 0.1758 0.0539} 
\definecolor{myred}{rgb}{0.4758, 0.0328, 0.0539}
\definecolor{myred2}{rgb}{0.75, 0.0328, 0.0539}

\title{Continual Learning with Distributed Optimization: \\
Does COCOA Forget? }

\begin{document}
    
    \author{\IEEEauthorblockN{Martin Hellkvist, Ay\c ca \"Oz\c celikkale, Anders Ahl\'{e}n}
    \IEEEauthorblockA{\textit{Department of Electrical Engineering,}
    Uppsala University, Sweden \\
    \{Martin.Hellkvist, Ayca.Ozcelikkale, Anders.Ahlen\}@angstrom.uu.se}
    }
    \maketitle
    
    \begin{abstract}
    We focus on the continual learning problem where the tasks arrive sequentially and the aim is to perform well on the newly arrived task without performance degradation on the previously seen tasks.  
    In contrast to the continual learning literature focusing on the centralized setting, we investigate the distributed estimation framework. We consider the well-established distributed learning algorithm \cocoa{}. We derive closed form expressions for the iterations for the overparametrized case. We illustrate the convergence and the error performance of the algorithm based on the over/under-parametrization of the problem.  
    Our results show that depending on the problem dimensions and data generation assumptions, 
    \cocoa{} can perform continual learning over a sequence of tasks,
    i.e.,
    it can learn a new task without forgetting previously learned tasks,
    with access only to one task at a time.
\end{abstract}

    \newcommand{\mycontinualcocoa}{
    
    \SetKw{KwEnd}{end}
	\textbf{Input}: Tasks 
	$\Sc\!=\!\{\!\tasktau{t}\!\}_{t=1}^T$,
	partitioning 
	scheme 
	$\!\!\Pc\!=\!\{p_k\}_{k=1}^K$,
    number of iterations $T_c$ for \cocoa{} to run per task.
    
    \textbf{Initialize}: $\wvec_0 = \bm 0$.
	
	\For{$t = 1,\,\dots,\,T$}{
	
    $\xvec^0=\wvec_{t-1}$
    
    $\vvec_k^0{=} K \Atauk{t}{k} \wvec_{t-1,[k]} \,\, \forall\, k$.
    
    \setstretch{1.05}
    \For{$i=0,\,1,\,\dots,\,T_c - 1$}{
        $\vbar^i = \frac{1}{K}\sum_{k=1}^K\vvec_k^i$
        
        \For{$k \in \{1,\,2,\,\dots,\,K\}$}{
        
            $\dxki = \frac{1}{K}\Atauk{t}{k}\p\left(\ytau{t} - \vbar^i\right)$
            
            $\xvec_{[k]}^{i+1} = \xvec_{[k]}^i + \dxki$
            
            $\vvec_k^{i+1} = \vbar^i + K\Atauk{t}{k}\dxki$
        }
    }
    $\wvec_t = \xvec^{T_c}$
	}
    \caption{Implementation of \cocoa{} \cite{smith_cocoa_nodate} for \eqref{eqn:cocoa_opt_fun}.%
    \label{alg:continual_cocoa} 
    }
    }
    \newtheorem{thm}{\bf{Theorem}}
\newtheorem{cor}{\bf{Corollary}}
\newtheorem{lem}{\bf{Lemma}}
\newtheorem{prop}{\bf{Proposition}}
\newtheorem{rem}{\bf{Remark}}

\theoremstyle{remark} 
\newtheorem{defn}{\bf{Definition}}[section]
\newtheorem{ex}{\bf{Example}}[section]
\newtheorem{asmptn}{\bf{Assumption}}
\newtheorem{myexp}{\bf{Experiment}}

\newenvironment{theorem}
{\par\noindent \thm \begin{itshape}\noindent}
{\end{itshape} \vspace{3pt}}

\newenvironment{lemma}
{\par\noindent  \lem \begin{itshape}\noindent}
{\end{itshape} \vspace{3pt}}

\newenvironment{corollary}
{\par\noindent  \cor \begin{itshape}\noindent}
{\end{itshape} \vspace{3pt}
}

\newenvironment{remark}
{\par\noindent \rem \begin{itshape}\noindent}
{\end{itshape} \vspace{3pt}}

\newenvironment{definition}
{\par\noindent \defn \begin{itshape}\noindent}
{\end{itshape}}

\newenvironment{assumption}
{\par\noindent \asmptn \begin{itshape}\noindent}
{\end{itshape}}

\newenvironment{experiment}
{\vspace{3pt} \par\noindent \myexp \begin{itshape} \noindent}
{\end{itshape}\vspace{6pt}}

\newenvironment{example}
{\vspace{2pt} \par\noindent \ex} 
{\vspace{2pt}}

    \section{Introduction}\label{sec:introduction}
% \kern-0.5em
When presented with a stream of data,
continual learning \cite{Parisi_continual_2019, Kirkpatrick_catastrophic_2017} is the act of learning from new data while not forgetting what was learnt previously.
New data can, for instance, come from a related classification task with new fine-grained classes,
or it can have statistical distribution shift compared to the previously seen data. 
Each set of data that is presented to the model is referred to as a \textit{task}. 
Continual learning aims to create models which perform well on all seen tasks without the need to retrain from scratch when new data comes  \cite{Kirkpatrick_catastrophic_2017, Parisi_continual_2019, Evron_catastrophic_2022, French_Catstrophic_1999}.

The central issue in continual learning is \textit{forgetting}, 
which measures the performance degradation on previously learned tasks as new tasks are learnt by the model \cite{French_Catstrophic_1999, Evron_catastrophic_2022, Kirkpatrick_catastrophic_2017}, see \eqref{eqn:forgetting_Defintion} in Section~\ref{sec:problemFormulation:task} for a formal definition. 
If a model performs poorly on old tasks, it is said to exhibit \textit{catastrophic forgetting} \cite{French_Catstrophic_1999, Evron_catastrophic_2022, Kirkpatrick_catastrophic_2017}.
To achieve continual learning,
the forgetting must be reduced.
Nevertheless, the  key enabling factors for achieving continual learning are not well established,
even for linear models  \cite{Kirkpatrick_catastrophic_2017, Evron_catastrophic_2022}.

Forgetting is closely related to the error 
performance
under non-stationary distributions. 
Various phenomena of interest, 
such as financial time-series and target tracking,
often exhibit structural changes in signal characteristics over time. Hence, performance under non-stationary distributions has been studied in a number of scenarios; 
including with random drift in the unknown within a distributed learning setting \cite{NosratiShamsiTaheriSedaaghi_2015, Sayed_Diffusion_LMS_2015} 
as well under switching system dynamics \cite{FoxSudderthJordanWillsky_2011,
DingShahrampirHealTarokh_2018,
KarimiButalZhaoKamalabadi_2022}.   

We consider the continual learning problem from a distributed learning perspective, where optimization is performed over a network of computational nodes.  
In addition to supporting scalability, distributed learning is  also attractive for scenarios where the data is already distributed over a network,
for example in sensor networks \cite{rabbat_distributed_2004, Kar_Distributed_2009,HuaNassifRicharWangSayed_2020,sayed2014adaptation}, or
in dictionary learning where sub-dictionaries are naturally separated over the network
\cite{Sayed_Dictionary_distributed_2015},
or in multi-task settings where the nodes have separate but related tasks 
\cite{Skoglund_2021_RNN_Multitask, Sayed_2014_multitask_networks}.

In this work, 
we focus on the well-established distributed learning algorithm \cocoa{}~\cite{smith_cocoa_nodate, jaggi2014communication, ma2017distributed, he_cola_2019, hellkvist_ozcelikkale_ahlen_linear_2021},
which allows the nodes to use a local solver of their choice for their local subproblems. 
In contrast to the setting of \cite{ Skoglund_2021_RNN_Multitask, Sayed_2014_multitask_networks} where the individual nodes have different tasks but these tasks do not change over time, 
here all nodes have the same task and this task changes over time for all nodes. 
We investigate the ability of
\cocoa{} 
to perform continual learning for a linear model.
We derive closed form expressions for the iterations of \cocoa{} when the number of unknown parameters at each node exceeds the number of samples per task.
We illustrate the performance of the algorithm with varying number of tasks and data points, 
revealing the trade-offs between the forgetting,
the convergence performance
and the dimensions of both the offline centralized problem,
i.e., the problem of solving all tasks in one large batch in a centralized manner,
and the nodes' local problems. 
{\bf{Our main findings}} can be summarized as follows:
\begin{itemize}[align=left, style=nextline, leftmargin=*, topsep=1pt, itemsep=0pt]
    \item  \cocoa{} can exhibit continual learning by modifying the initialization point of the algorithm.
    
    \item \cocoa{} does not forget if all the tasks are underparametrized and have a shared solution. 

    \item Even when each task comes only once,
    \cocoa{} can learn the new tasks with a reasonable forgetting performance. 
    Whether the convergence and forgetting can be improved with task repetition depends on the data generation model. 

    \item If the total number of samples over all tasks is close to the total number of parameters,
    then the algorithm may have slow convergence and relatively large error, 
    even when there is a shared solution that solves all the tasks.
    
\end{itemize}

    \section{Problem Statement}
\label{sec:problem}
% \kern-0.25em

\subsection{Definition of a Task}\label{sec:problemFormulation:task}
% \kern-0.25em
The data for task $\tau(t)$ consists of the feature matrix $\Atau{t}\!\in\!\Rbb^{\ntau{t} \!\times p}$,
and the corresponding vector of outputs
$\ytau{t} \inrbb{\ntau{t}\times 1}$.
We focus on fitting a linear model 
such that,
\begin{equation}\label{eqn:model}
    \ytau{t} = \Atau{t} \wvec_{\tau(t)}
\end{equation}
where $\wvec_{\tau(t)} \inrbb{p\times 1}$ is the vector of unknown model parameters.
We refer to the above regression problem with the data $\tasktau{t}$ as \textit{task} $\tau(t)$.

\subsection{Continual Learning}
We would like to fit the linear model in \eqref{eqn:model}
to a sequence of tasks,
\begin{equation}
    \Sc = \left\{\tasktau{t}\right\}_{t=1}^T. 
\end{equation}
At a given time instant $t$, we have  access to $\tasktau{t}$ but not to the other tasks.
Hence, we would like to solve all tasks in  $\Sc$ when we have access to only one of them at a given time.
Throughout the paper,
we apply the following assumption on the tasks $\tasktau{t}$,

\begin{assumption}\label{asmp:exists_soln}
    There exists a solution $\wstar\inrbb{p\times 1}$ that solves all the tasks in $\Sc$,
    i.e., 
    \begin{equation}\label{eqn:w_star_solves_all}
        \exists\, \wstar\inrbb{p\times 1}:\, \Atau{t} \wstar = \ytau{t},\,t=1,\,\dots,\, T.
    \end{equation}

\end{assumption}

\noindent
Note that Assumption~\ref{asmp:exists_soln} can be interpreted as a stationarity assumption.
This constitutes a reasonable scenario in the case of highly overparameterized models,
i.e., models that have more tuneable parameters than data samples,
and has been used in the centralized continual learning setting of \cite{Evron_catastrophic_2022}. 
We consider the investigation of this setting as good starting point for the general case since it facilitates a tractable analysis. 
However, it is not always natural to make this assumption, 
hence we 
study a setting where Assumption~\ref{asmp:exists_soln} does not hold in 
Section~\ref{sec:family_of_tasks}.

For a given parameter vector $\wvec$,
we measure the forgetting by the squared loss on a specific task $\tasktau{t}$,
i.e.,
$\frac{1}{\ntau{t}}\!\left\|\Atau{t} \wvec - \ytau{t}\right\|^2\!\!,$
where $\|\!\cdot\!\|$ denotes the $\ell_2$-norm,
as done in previous works \cite{Doan_Catastrophic_2021,Evron_catastrophic_2022}.
While training over the sequence of tasks $\Sc$,
a corresponding sequence of parameter vectors estimates is generated,
denoted by $\wvec_t,\,t\!=\!1,\dots,T$.
We denote the forgetting of $\wvec_t$
as the forgetting of the tasks up until and including the current task,
\begin{align}\label{eqn:forgetting_Defintion}
    \Fc_{\Sc} \left(t \right) 
    =  \frac{1}{t}
    \sum_{i=1}^{t} \frac{1}{\ntau{i}}
    \left\| \Atau{i} \wvec_t - \ytau{i} \right\|^2 .
\end{align}

Note that forgetting is defined in terms of data fit for $\ytau{i}$ instead of estimation accuracy for $\wstar$.

% \kern-1em
\subsection{Distributed Continual Learning with \cocoa{}}
% \kern-0.5em
\setlength{\textfloatsep}{0pt}
        \begin{algorithm}[t]
            \mycontinualcocoa
        \end{algorithm}

In \cocoa{}, the unknowns  are distributed over a network of $K$ nodes \cite{smith_cocoa_nodate, jaggi2014communication, ma2017distributed, he_cola_2019}. 
The unknown vector $\wvec$ is partitioned according to the partitioning $\Pc=\{p_k\}_{k=1}^K$,
such that node $k$ governs $p_k$ of the $p$ unknowns in $\wvec\inrbb{p\times 1}$. 
Note that this is in contrast to settings where the observations are distributed over the network. 
The partitioning consists of mutually exclusive sets of indices, hence
each entry of the unknown vector 
(hence each column of a given feature matrix)
is associated with only one node,
i.e., $\sum_{k=1}^K p_k = p$.

% %
For each task $\tasktau{t}$,
the columns of $\Atau{t}\inrbb{\ntau{t} \times p }$  is partitioned according to the partitioning for the unknowns.
We denote the matrix of columns associated with node $k$ as 
$\Atauk{t}{k}$.
For each presented task $\tau(t)$ in the sequence $\Sc$, 
we run \cocoa{} to minimize the forgetting of the current task,
i.e.,
$
    \min_{\wvec_t} \!
    \frac{1}{2 \ntau{t}}\left\| \Atau{t}\wvec_t \! - \! \ytau{t} \right\|^2 \!\!\! .
$
\cocoa{} is an iterative algorithm,
which is run for $T_c$ iterations for each task.
For notational clarity, the unknown vector is represented by $\wvec_t$ for the outer iteration $t$ over the tasks and by $\xvec_i$ for the inner \cocoa{} iteration $i$. 
The partitionings of $\wvec_t$ and $\xvec_i$ which correspond to the local matrices $\Atauk{t}{k}$ are denoted by
$\wvec_{t,[k]}$ and $\xvec_{i,[k]}$,
respectively.
Hence the overall training scheme,
presented in Algorithm~\ref{alg:continual_cocoa},
consists of three nested for-loops:
the \textit{outer iterations} $t=1,\,\dots,\,T$ over the sequence of tasks;
the \textit{inner iterations} $i=0,\,\dots,\, T_c-1$ of \cocoa{};
and the parallelization over the nodes $k=1,\,\dots,\,K$.

For task $\tau(t)$,
the nodes iteratively update their local variables $\xvec_{i,[k]}$ by finding the update $\dxki$
as the solution to the following local subproblem in iteration $i=0,1,\dots,T_c-1$,
\cite[Sec.~3.1]{smith_cocoa_nodate}
\begin{align}\label{eqn:cocoa_opt_fun}
\begin{split}
    \min_{\dxki} 
    & \frac{1}{2K \ntau{t}} \! \left\| \vbar^i - \ytau{t} \right\|^2 
     \! \! + \! \!  \frac{\subproblemp}{2\ntau{t}}\left\| \Atauk{t}{k} \dxki \right\|^2 \\
    & \qquad + \frac{1}{\ntau{t}}\left(\vbar^i - \ytau{t}\right)\T \Atauk{t}{k}\dxki,
\end{split}
\end{align}
where $\vbar^i$ is the aggregated shared estimation of $\ytau{t}$ in iteration $i$ and Here, $\subproblemp$ is a hyperparameter for the algorithm.
This is a convex problem in $\dxki$. 
Setting the gradient w.r.t. $\dxki$ to zero
and choosing the minimum $\ell_2$-norm solution, we obtain the solution as 
% %
$
    \dxki = \frac{1}{\subproblemp} \Atauk{t}{k}\p \left(\ytau{t} - \vbar^i\right),
$
where $(\cdot)\p$ denotes the Moore-Penrose pseudoinverse.
Using the notation of \cite{smith_cocoa_nodate},
we set  $\subproblemp=\aggregationp K$,
with $\aggregationp \in (0, 1]$,
as these are considered safe choices \cite{he_cola_2019}.
% %
With these choices, these parameters cancel out to give the explicit expressions in Algorithm~\ref{alg:continual_cocoa}.

In order to utilize the knowledge from the previous tasks, we do the following: 
i) we initialize \cocoa{} with the parameter vector from the previous step,
i.e., we set $\xvec^0 = \wvec_{t-1}$; 
ii) we set the initial point for the local contributions as 
$\vvec_k^0 = K \Atauk{t}{k} \xvec^0_{[k]}$.

\subsection{Offline and centralized problem}
Throughout our analysis of the distributed continual learning problem,
we consider the \textit{offline and centralized problem} as the reference problem,
which corresponds to solving the unique set of tasks in $\Sc$ simultaneously in a centralized fashion,
i.e.,
\begin{equation}\label{eqn:simultaneous_problem}
    \!\!\!
    \Yvec \! = \! \Abb \wvec,\,
    \Yvec \! = \!\!\begin{bmatrix}
        \yvec_1 \! \\
        \vdots  \!\\
        \yvec_M \!
    \end{bmatrix}\!\!\!\inrbb{N\times 1}, \,
    \Abb = \!
    \begin{bmatrix}
        \Amat_1 \\
        \vdots \\   
        \Amat_M
    \end{bmatrix}\!\!\inrbb{N\times p},
\end{equation}
where $\{ \task{m}\}_{m=1}^M$ is the unique set of tasks in $\Sc$,
with $\Amat_m\inrbb{n_m\times p}$ and $\yvec_m \inrbb{n_m\times 1}$,
and the total number of samples over all the unique tasks is 
$
    N=\sum_{m=1}^M n_m.
$
Throughout the paper,  we  use subindices $\tau(t)$ and $m$ to emphasize different task settings.
The notation $\{ \task{m}\}_{m=1}^M$ refers to the unique set of $M$ tasks in $\Sc$,
and is used when each task appears once. 
The notation $\tau(t)\in\Nbb_+$, $t=1,\,\dots,\,T$,
is used when tasks can be repeated.

\section{Overparameterized Local Problems}
If $p_k \geq \ntau{t}$ for all $k=1,\,\dots,\,K$ and $t=1,\,\dots,\,T$,
then the local problems are overparameterized,
i.e., the nodes' local feature matrices $\Atauk{t}{k}\inrbb{\ntau{t}\times p_k}$ are broad.
In this section,
we assume that the matrices $\Atauk{t}{k}$ have full rank,
i.e., $\ntau{t}$,
thus
\begin{equation}\label{eqn:A\p_times_A_identity}
     \Atauk{t}{k}\Atauk{t}{k}\p = \eye{\ntau{t}}.
\end{equation}
Using \eqref{eqn:A\p_times_A_identity},  
the updates of $\vvec_k^{i+1}$ for $i\geq 0$ 
(line 11 in Algorithm~\ref{alg:continual_cocoa}) 
can 
be rewritten as
\begin{equation}
    \vvec_k^{i+1}
    = \vbar^i + \Atauk{t}{k}\Atauk{t}{k}\p \left(\ytau{t} - \vbar^i\right)
    = \ytau{t}.
\end{equation}
It follows that
$
    \vbar^i = \frac{1}{K}\sum_{k=1}^K \vvec_k^{i} = \ytau{t},~i\geq 1
$
and as a result,
the steps $\dxki$ for $i\geq 1$ are zero,
i.e.,
\begin{align}
\begin{split}
    \dxki 
    & = \frac{1}{K}\Atauk{t}{k}\p\left(\ytau{t} - \vbar^i\right) 
    = \bm 0.
\end{split}
\end{align}
In other words,
\cocoa{} in this setting converges after its first iteration $i=0$.
To find the solution $\xvec^1$ that \cocoa{} converges to,
we insert the initial values of
$\xvec^0 = \wvec_{t-1}$,
and $\vvec_k^0 = K \Atauk{t}{k}\wvec_{t-1,[k]}$,
finding that 
\begin{equation}
    \vbar^0 = \Atau{t}\wvec_{t-1},
\end{equation}
and that the local estimates $\xvec_k^i$ in $\cocoa{}$ converge to
\begin{align}
    \xvec_{[k]}^i
    & = \wvec_{t-1,[k]} + \frac{1}{K}\Atauk{t}{k}\p \left( \ytau{t} - \Atau{t}\wvec_{t-1} \right).
\end{align}
Combining over the nodes $k=1,\dots,K$,
the parameter vector $\wvec_t = \xvec^1$ after training on task $\tau(t)$ is,
\begin{equation}\label{eqn:wvec_t_overparameterized_task}
    \wvec_t = \Ptau{t} \wvec_{t-1} + \Abartau{t}\ytau{t},
\end{equation}
where we have introduced the notation
\begin{align}
    \Ptau{t} &= \left(\eye{p} - \Abartau{t}\Atau{t}\right), \\
    \Abartau{t}\! & =\! \frac{1}{K}\!
        \left[(\Atauk{t}{1}\p)\T, \cdots,
        (\Atauk{t}{K}\p)\T\right]\T \!\!\!\! \inrbb{p\times \ntau{t}}.
\end{align}
In this setting,
\cocoa{} always solves the latest seen task,
i.e.,
\begin{align}
    \hat \yvec_{\tau(t)} 
    & = \Atau{t} \wvec_t 
    \\
    & 
    = \Atau{t} \Ptau{t} \wvec_{t-1} + \Atau{t}\Abartau{t}\ytau{t} \\
    & = \left(\Atau{t} - \Atau{t}\right)\wvec_{t-1} + \ytau{t}
    = \ytau{t},
\end{align}
where we have used that 
$\Atau{t}\Abartau{t}\!  = \! \eye{\ntau{t}} $
as a result of \eqref{eqn:A\p_times_A_identity}.
These findings are summarized in the following lemma.
\begin{lemma}\label{remark:one_step_cocoa}
    If the partitions $\Atauk{t}{k}\inrbb{\ntau{t}\times p_k}$ 
    are full rank,
    and $p_k \geq \ntau{t}$ for all $\krange$,
    then \cocoa{} (lines 4 -- 11 in Algorithm~\ref{alg:continual_cocoa})
    converges after the first iteration $i=0$ when run on the task $\tasktau{t}$,
    and
    the resulting solution $\wvec_t$ in \eqref{eqn:wvec_t_overparameterized_task} solves the task,
    i.e., $\hat{\yvec}_{\tau(t)} =  \Atau{t}\wvec_t = \ytau{t}$.
\end{lemma}

Although the algorithm after each iteration $t$
always solves the latest task, 
these solutions do not necessarily perform well on previously seen tasks.
We now continue with investigating the convergence and forgetting dynamics of $\wvec_t$ in \eqref{eqn:wvec_t_overparameterized_task}.

\subsection{One shot at each task}\label{section:one_shot_training}

\begin{figure}[t]
    \centering
    \includegraphics[width=0.8\linewidth]{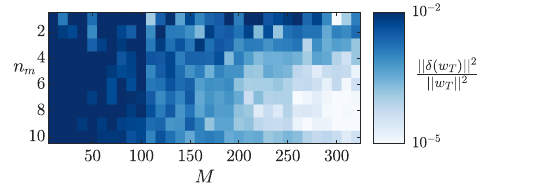}
    \caption{The size of the final step from $\wvec_{T-1}$ to $\wvec_T$ after training on each task once,
    versus the number of tasks $M$ and the number of samples per task $n_m$.}
    \label{fig:convergence_one_shot}
\end{figure}
\begin{figure}[t]
    \centering
    \includegraphics[width=0.8\linewidth]{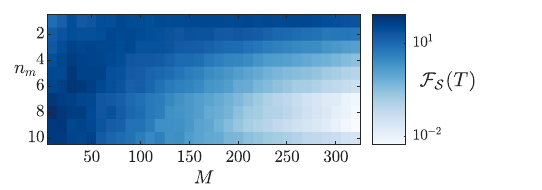}
    \caption{The forgetting of $\wvec_T$ after training on each task once, 
    versus the number of tasks $M$ and the number of samples per task $n_m$.}
    \label{fig:forgetting_one_shot}
\end{figure}

\begin{figure}[t]
    \centering
    \includegraphics[width=0.8\linewidth]{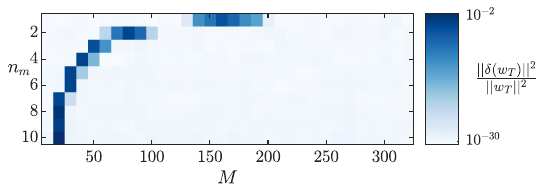}
    \caption{The size of the final step from $\wvec_{T-1}$ to $\wvec_T$ after training on each task 1000 times,
    versus the number of tasks $M$ and the number of samples per task $n_m$.}
    \label{fig:convergence_cyclic}
\end{figure}
\begin{figure}[t]
    \centering
    \includegraphics[width=0.8\linewidth]{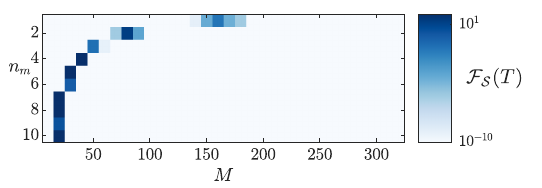}
    \caption{The forgetting of $\wvec_T$ after training on each task 1000 times, 
    versus the number of tasks $M$ and the number of samples per task $n_m$.}
    \label{fig:forgetting_cyclic}
\end{figure}

We now consider a setting where every task only comes once in the sequence $\Sc$.
In this section,
we discuss numerical results from simulations
where we generate the tasks as
$\yvec_m = \Amat_m\wstar$, $m=1,\dots,M$,
where $\Amat_m\inrbb{n_m\times p}$ have independently generated standard Gaussian entries.
We fix the number of parameters $p=160$ and vary the number of samples per task $n_m$.
We set the generating solution to $\wstar = \bm 1$.
We set the training sequence as $\tau(t) = t$, $t=1,\,\dots,\, T$ with $T=M$,
thus we train the model on each task once.
We measure the convergence as the relative size of the last step taken by the outer iterations, i.e.,
$
    \frac{\|\deltavec(\wvec_T)\|^2}{\|\wvec_T\|^2},
$
with $\deltavec(\wvec_T) = \|\wvec_T - \wvec_{T-1}\|^2$.
We fix the number of nodes in the network as $K=4$,
which together perform the \cocoa{} iterations,
and the partitioning as $\Pc=\{p_k\}_{k=1}^K = \{16, 32, 48, 64\}$.
Note that we only study $n_m \leq 10$,
since if $n_m \approx p_k$ for any local problem,
then \cocoa{} obtains extremely high error values with high probability \cite{hellkvist_ozcelikkale_ahlen_linear_2021}.
Note that the task feature matrices $\Amat_m$ considered here are generated as standard Gaussian matrices,
hence their partitions $\Amat_{m,[k]}$
are full rank with probability 1, 
and the tasks fulfill Lemma~\ref{remark:one_step_cocoa}.
We plot the convergence in Figure~\ref{fig:convergence_one_shot}
and the forgetting in the final step $\Fc_{\Sc}(T)$ in Figure~\ref{fig:forgetting_one_shot}. 
We observe the following:

\begin{remark}
    Even though \cocoa{},
    i.e., the inner iterations of Algorithm~\ref{alg:continual_cocoa},
    converge in one step and solves the latest seen task
    (see Lemma~\ref{remark:one_step_cocoa}),
    the obtained solution $\wvec_T$ after seeing all tasks once 
    can exhibit forgetting if  the number of tasks $M$ or the number of data points per task $n_m$  is too small.
    The forgetting can be improved by increasing the number of tasks $M$ (hence increasing the number of data points) due to existence of the solution $\wstar$ that solves all tasks. 
\end{remark}

\subsection{Cyclic sequence of tasks}\label{section:cyclic_training}

We now repeat the experiment of Section~\ref{section:one_shot_training},
but extend the training sequence so that the model trains on each task $1000$ times.
In other words,
with $M$ tasks we set the sequence as
$\tau(t) = 1,\,\dots,\,M,\,1,\,\dots,\,M,\dots$, $t=1,\,\dots,\,T$,
with $T=1000 M$.
We plot the convergence in Figure~\ref{fig:convergence_cyclic} and the forgetting in Figure~\ref{fig:forgetting_cyclic}.
The plots illustrate that the algorithm converges for most pairs of $(n_m,\,M)$,
and that the forgetting for the final solution is low.
However,
the plots also illustrate that both the convergence and forgetting are worse along a streak of pairs $(n_m, M)$,
for which $N \approx p$,
i.e., where the total number of samples $N = \sum_{m=1}^M n_m $ is close to or equal to the number of parameters $p$,
for example $(1,160)$, $(2,80)$ and $(4,40)$.
We now analyze this phenomenon,
distinguishing between the following scenarios:
\textit{i)} $N > p$;
\textit{ii)} $N \approx p$;
and \textit{iii)} $N < p$. 
In the below, recall that $n_m$ is constant over $m=1,\,\dots,\,M$,
hence $N = M n_m$.
Recall that we generate the matrices $\Amat_m$ as independent standard Gaussian matrices,
and as a result, 
the offline and centralized matrix $\Abb$ has full rank with probability one.

\textit{i)} $N > p$:
If $M$ or $n_m$ is large enough so that
the total number of samples $N = M n_m$ exceeds the number of parameters $p$,
then the offline and centralized problem $ \Yvec = \Abb\wvec $ in \eqref{eqn:simultaneous_problem}
is underparameterized,
i.e., there are less parameters than samples.
The matrix $\Abb\inrbb{N\times p}$ has full rank,
and  $\Yvec = \Abb\wstar$. 
Hence, $\wstar$ is the unique solution to the offline and centralized problem.
This suggests that $\wvec_T = \wstar$, 
since the solutions for large $M$ or $n_m$ in Figure~\ref{fig:convergence_cyclic} and \ref{fig:forgetting_cyclic} converge to a solution $\wvec_T$ which solves all tasks,
and $\wstar$ is the only solution which does so.

\begin{figure}
    \centering
    \includegraphics[width=0.8 \linewidth]{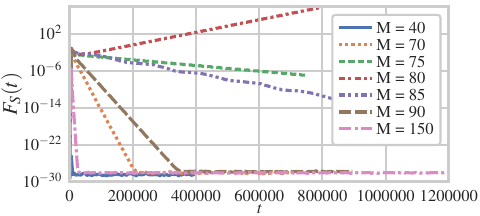}
    \caption{The forgetting of $\wvec_t$ versus $t$ for different number of tasks $M$ with $n_m=2$ samples each.}
    \label{fig:t_forgetting_cyclic}
\end{figure}

\textit{ii) } $N\approx p$:
If the total number of samples $N = M n_m$ is close to or equal to the number of parameters $p$,
then the offline and centralized feature matrix $\Abb$ is nearly square or square.
The results in Figure~\ref{fig:convergence_cyclic} and \ref{fig:forgetting_cyclic} illustrate that Algorithm~\ref{alg:continual_cocoa} does not converge,
and does not solve the previously seen tasks.
The non-convergence of the algorithm for these pairs of $(n_m, M)$ are further illustrated in Figure~\ref{fig:t_forgetting_cyclic},
where we plot the forgetting as a function of the outer iterations $t$ in Algorithm~\ref{alg:continual_cocoa},
i.e., $\Fc_{\Sc}(t)$,
where we have evaluated the forgetting every $1000$ iterations,
and fixed the number of samples per task to $n_m = 2$.
Figure~\ref{fig:t_forgetting_cyclic} illustrates that 
if $M=80$,
then the vectors $\wvec_t$ do not converge,
and are instead diverging.
If $M\approx 80$ but $M\neq 80$,
then the convergence is very slow compared to the case with $M$ further away.

\begin{figure}
    \centering
    \includegraphics[width=1 \linewidth]{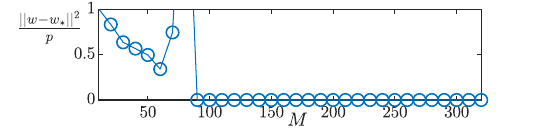}
    \caption{The distance from $\wvec_T$ to the solution $\wstar=\bm 1$ used to generate the data $y_m$.
    Here, $n_m = 2$ for each task.}
    \label{fig:distance_to_wstar}
\end{figure}

\textit{iii)} $N < p$: 
In this scenario,
the offline and centralized problem is overparameterized.
Hence, there is possibly an infinite number of solutions $\wvec$ which solve all the tasks simultaneously.
In this regime,
$\wvec_t$ generally do not converge to  $\wstar$.
In Figure~\ref{fig:distance_to_wstar},
we plot the distance from $\wvec_T$ to the solution $\wstar$ 
versus the number of tasks $M$ for the experiment in Figure~\ref{fig:convergence_cyclic} and \ref{fig:forgetting_cyclic}.
Figure~\ref{fig:distance_to_wstar} illustrates that for small $M$,
even though the final solution $\wvec_T$ solves all the tasks,
it is not equal to $\wstar = \bm 1$.
Consistent with our observations for scenario \textit{i)},
Figure~\ref{fig:distance_to_wstar} suggests that $\wvec_t$ converges to $\wstar$ for large $M$.

\section{Underparameterized Task} 
\label{section:one_underparameterized}
A task $\tasktau{t}$ where $\Atau{t}\inrbb{\ntau{t}\times p}$ is such that 
$p < \ntau{t}$ is referred to as \textit{underparameterized},
as there is a smaller number of tuneable parameters in $\wvec\inrbb{p\times 1}$
than the number of equations in $\Atau{t}\wvec = \ytau{t}$.
If the feature matrix $\Atau{t}$ of an underparameterized task is full rank,
then $\wstar$ from Assumption~\ref{asmp:exists_soln}
is the unique solution to that task.
By \cite[Theorem~2]{smith_cocoa_nodate},
a sufficiently large number of iterations $T_c$ of \cocoa{} guarantees
an arbitrarily small optimality gap.
In other words,
if $T_c$ is large enough while learning an underparameterized task $\tasktau{t}$,
then the estimate $\xvec^{T_c}$ of \cocoa{} will converge to $\wstar$.
Hence, we have the following lemma:
\begin{lemma}
    If all tasks in the sequence $\Sc$ are underparameterized and have full rank feature matrices,
    and $T_c\rightarrow \infty$,
    then Algorithm~\ref{alg:continual_cocoa} converges to $\wstar$,
    which is the unique solver of all tasks in $\Sc$.

\end{lemma}

\section{Family of Tasks without Assumption 1}
\label{sec:family_of_tasks}

\begin{figure}
    \centering
    \includegraphics{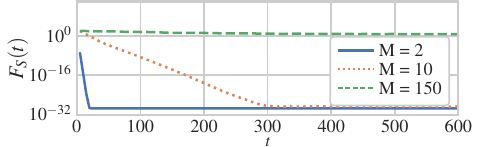}
    \caption{The forgetting of $\wvec_t$ versus $t$ when the task output vectors $\yvec_m$ are generated either by $\weven$ or $\wodd$.}
    \label{fig:forgetting_alternating_tasks}
\end{figure}

We now consider a setting where Assumption~\ref{asmp:exists_soln} does not  hold.
We again consider the cyclic manner of training on $M$ tasks as in Section~\ref{section:cyclic_training}.
Tasks $\task{m}$ with even and odd indices $m$ are generated with $\yvec_m = \Amat_{m} \weven$,
and $\yvec_m = \Amat_{m}\wodd$,
respectively.
We set $\weven = \bm 1_{p\times 1}$,
and $\wodd = [\bm 1_{1\times 0.9 p }, \bm{0}_{1\times0.1p}]\T$. 
We randomly create the 
features matrices $\Amat_m$ as independent standard Gaussian matrices. We set $n_m = 2$ samples  and vary the number of tasks $M$.

In Figure~\ref{fig:forgetting_alternating_tasks},
we present the forgetting $\Fc_{\Sc}(t)$ versus the number of outer iterations $t$ of Algorithm~\ref{alg:continual_cocoa},
for different values of $M$.
These plots illustrate the different convergence behaviour  for small and large $M$.  In particular, 
for small values of $M$ ($M=2$),
the algorithm quickly converges to a solution which solves all the tasks. 
(Although $\weven \neq \wodd$, there can be such a shared solution since the centralized, offline system is 
overparameterized
for $M$ small.)
For larger values of $M$ ($M=10$), convergence is slower. For higher values of $M$ ($M=150$),   
the algorithm does not converge.
Although not visible in Figure~\ref{fig:forgetting_alternating_tasks},
the forgetting for $M=150$ fluctuates on the range $[4,7]$ in the plot for $t>2000.$

These results illustrate that 
although the total number of samples for the two types of task with $\weven$ and $\wodd$ increases with $M$, 
the forgetting does not necessarily improve. This is consistent with the fact that for larger values of $M$ there is a higher number of equations that needs to be satisfied by the  solution vector of \cocoa{}.

    \section{Conclusions}\label{sec:conclusions}
We have applied the distributed learning algorithm \cocoa{} in a continual learning setting,
and investigated its properties in terms of convergence and forgetting.
Our results illustrate that \cocoa{} can be suitable for performing continual learning in a distributed manner,
and that the forgetting can be reasonably low even if each task only comes once in the sequence. We have illustrated how 
the dynamics of the convergence and forgetting of the algorithm 
vary with the dimensions of both the local problems at the nodes as well as the offline and centralized problem.
There are several interesting lines of future work, including characterization of convergence and forgetting behaviour under other data models as well as development of methods to improve the forgetting behaviour of the algorithm.

    \kern-0.25em
    
    \bibliographystyle{IEEEtran}
    \bibliography{ref}    

% Generated by IEEEtran.bst, version: 1.12 (2007/01/11)
\begin{thebibliography}{10}
\providecommand{\url}[1]{#1}
\csname url@samestyle\endcsname
\providecommand{\newblock}{\relax}
\providecommand{\bibinfo}[2]{#2}
\providecommand{\BIBentrySTDinterwordspacing}{\spaceskip=0pt\relax}
\providecommand{\BIBentryALTinterwordstretchfactor}{4}
\providecommand{\BIBentryALTinterwordspacing}{\spaceskip=\fontdimen2\font plus
\BIBentryALTinterwordstretchfactor\fontdimen3\font minus \fontdimen4\font\relax}
\providecommand{\BIBforeignlanguage}[2]{{%
\expandafter\ifx\csname l@#1\endcsname\relax
\typeout{** WARNING: IEEEtran.bst: No hyphenation pattern has been}%
\typeout{** loaded for the language `#1'. Using the pattern for}%
\typeout{** the default language instead.}%
\else
\language=\csname l@#1\endcsname
\fi
#2}}
\providecommand{\BIBdecl}{\relax}
\BIBdecl

\bibitem{Parisi_continual_2019}
G.~I. Parisi, R.~Kemker, J.~L. Part, C.~Kanan, and S.~Wermter, ``Continual lifelong learning with neural networks: A review,'' \emph{Neural Networks}, vol. 113, pp. 54--71, 2019.

\bibitem{Kirkpatrick_catastrophic_2017}
J.~Kirkpatrick, R.~Pascanu, N.~Rabinowitz, J.~Veness, G.~Desjardins, A.~A. Rusu, K.~Milan, J.~Quan, T.~Ramalho, A.~Grabska-Barwinska, D.~Hassabis, C.~Clopath, D.~Kumaran, and R.~Hadsell, ``Overcoming catastrophic forgetting in neural networks,'' \emph{Proc. of the National Academy of Sciences}, vol. 114, no.~13, pp. 3521--3526, mar 2017.

\bibitem{Evron_catastrophic_2022}
I.~Evron, E.~Moroshko, R.~Ward, N.~Srebro, and D.~Soudry, ``How catastrophic can catastrophic forgetting be in linear regression?'' in \emph{Proc. of Thirty Fifth Conf. on Learning Theory}, vol. 178.\hskip 1em plus 0.5em minus 0.4em\relax PMLR, Jul 2022, pp. 4028--4079.

\bibitem{French_Catstrophic_1999}
R.~M. French, ``Catastrophic forgetting in connectionist networks,'' \emph{Trends in Cognitive Sciences}, vol.~3, no.~4, pp. 128--135, 1999.

\bibitem{NosratiShamsiTaheriSedaaghi_2015}
H.~Nosrati, M.~Shamsi, S.~M. Taheri, and M.~H. Sedaaghi, ``Adaptive networks under non-stationary conditions: Formulation, performance analysis, and application,'' \emph{IEEE Transactions on Signal Processing}, vol.~63, no.~16, pp. 4300--4314, 2015.

\bibitem{Sayed_Diffusion_LMS_2015}
J.~Chen, C.~Richard, and A.~H. Sayed, ``Diffusion lms over multitask networks,'' \emph{IEEE Transactions on Signal Processing}, vol.~63, no.~11, pp. 2733--2748, 2015.

\bibitem{FoxSudderthJordanWillsky_2011}
E.~Fox, E.~B. Sudderth, M.~I. Jordan, and A.~S. Willsky, ``Bayesian nonparametric inference of switching dynamic linear models,'' \emph{IEEE Transactions on Signal Processing}, vol.~59, no.~4, pp. 1569--1585, 2011.

\bibitem{DingShahrampirHealTarokh_2018}
J.~Ding, S.~Shahrampour, K.~Heal, and V.~Tarokh, ``Analysis of multistate autoregressive models,'' \emph{IEEE Transactions on Signal Processing}, vol.~66, no.~9, pp. 2429--2440, 2018.

\bibitem{KarimiButalZhaoKamalabadi_2022}
P.~Karimi, M.~D. Butala, Z.~Zhao, and F.~Kamalabadi, ``Efficient model selection in switching linear dynamic systems by graph clustering,'' \emph{IEEE Signal Processing Letters}, vol.~29, pp. 2482--2486, 2022.

\bibitem{rabbat_distributed_2004}
M.~Rabbat and R.~Nowak, ``Distributed optimization in sensor networks,'' \emph{Proc. 3rd Int. Symp. on Inf. Proc. in Sensor Netw.}, pp. 20--27, 2004.

\bibitem{Kar_Distributed_2009}
S.~Kar and J.~M.~F. Moura, ``Distributed consensus algorithms in sensor networks with imperfect communication: Link failures and channel noise,'' \emph{IEEE Transactions on Signal Processing}, vol.~57, no.~1, pp. 355--369, 2009.

\bibitem{HuaNassifRicharWangSayed_2020}
F.~Hua, R.~Nassif, C.~Richard, H.~Wang, and A.~H. Sayed, ``Diffusion {LMS} with communication delays: Stability and performance analysis,'' \emph{IEEE Signal Processing Letters}, vol.~27, pp. 730--734, 2020.

\bibitem{sayed2014adaptation}
A.~H. Sayed, ``Adaptation, learning, and optimization over networks,'' \emph{Foundations and Trends in Machine Learning}, vol.~7, pp. 311--801, 2014.

\bibitem{Sayed_Dictionary_distributed_2015}
J.~Chen, Z.~J. Towfic, and A.~H. Sayed, ``Dictionary learning over distributed models,'' \emph{IEEE Trans. on Signal Process.}, vol.~63, no.~4, pp. 1001--1016, 2015.

\bibitem{Skoglund_2021_RNN_Multitask}
Y.~Ye, M.~Xiao, and M.~Skoglund, ``Randomized neural networks based decentralized multi-task learning via hybrid multi-block {ADMM},'' \emph{IEEE Transactions on Signal Processing}, vol.~69, pp. 2844--2857, 2021.

\bibitem{Sayed_2014_multitask_networks}
J.~Chen, C.~Richard, and A.~H. Sayed, ``Multitask diffusion adaptation over networks,'' \emph{IEEE Transactions on Signal Processing}, vol.~62, no.~16, pp. 4129--4144, 2014.

\bibitem{smith_cocoa_nodate}
V.~Smith, S.~Forte, C.~Ma, M.~Tak{\'a}{\v{c}}, M.~I. Jordan, and M.~Jaggi, ``\textsc{CoCoA}: A general framework for communication-efficient distributed optimization,'' \emph{J. Mach. Learn. Res.}, vol.~18, no.~1, pp. 8590--8638, 2017.

\bibitem{jaggi2014communication}
M.~Jaggi, V.~Smith, M.~Tak{\'a}c, J.~Terhorst, S.~Krishnan, T.~Hofmann, and M.~I. Jordan, ``Communication-efficient distributed dual coordinate ascent,'' \emph{Adv. Neural Inf. Process. Systems}, pp. 3068--3076, 2014.

\bibitem{ma2017distributed}
C.~Ma, J.~Kone{\v{c}}n{\`y}, M.~Jaggi, V.~Smith, M.~I. Jordan, P.~Richt{\'a}rik, and M.~Tak{\'a}{\v{c}}, ``Distributed optimization with arbitrary local solvers,'' \emph{Optimization Methods and Softw.}, vol.~32, no.~4, pp. 813--848, 2017.

\bibitem{he_cola_2019}
L.~He, A.~Bian, and M.~Jaggi, ``Cola: Decentralized linear learning,'' \emph{Adv. Neural Inf. Process. Syst.}, pp. 4536--4546, 2018.

\bibitem{hellkvist_ozcelikkale_ahlen_linear_2021}
M.~Hellkvist, A.~Özçelikkale, and A.~Ahlén, ``Linear regression with distributed learning: A generalization error perspective,'' \emph{IEEE Trans. on Signal Process.}, vol.~69, pp. 5479--5495, 2021.

\bibitem{Doan_Catastrophic_2021}
T.~Doan, M.~Abbana~Bennani, B.~Mazoure, G.~Rabusseau, and P.~Alquier, ``A theoretical analysis of catastrophic forgetting through the ntk overlap matrix,'' ser. Proceedings of Machine Learning Research, A.~Banerjee and K.~Fukumizu, Eds., vol. 130, 2021, pp. 1072--1080.

\end{thebibliography}

\end{document}